\documentclass[10pt,letterpaper]{article}
\usepackage[top=0.85in,left=2.75in,footskip=0.75in]{geometry}

\usepackage{amsmath,amssymb}
\usepackage{changepage}
\usepackage{textcomp,marvosym}
\usepackage{cite}
\usepackage{nameref,hyperref}
\usepackage[nopatch=eqnum]{microtype}
\DisableLigatures[f]{encoding = *, family = * }
\usepackage[table]{xcolor}
\usepackage{array}
\usepackage{booktabs}
\usepackage{multirow}
\usepackage{boxedminipage}
\usepackage{tabularx}

\newcolumntype{+}{!{\vrule width 2pt}}

\newlength\savedwidth

\raggedright
\newlength{\extralength}
\usepackage[aboveskip=1pt,labelfont=bf,labelsep=period,justification=raggedright,singlelinecheck=off]{caption}

\makeatletter
\renewcommand{\@biblabel}[1]{\quad#1.}
\makeatother

\usepackage{lastpage,fancyhdr,graphicx}
\usepackage{epstopdf}
\fancyheadoffset[L]{\extralength}
\fancyfootoffset[L]{\extralength}
\newcommand{\tabref}[1]{Table~\ref{#1}}

\newcommand{\eg}{\textit{e.g.},}
\newcommand{\ie}{\textit{i.e.},}

\newcommand{\etal}{\textit{et al.}}

\begin{document}
\vspace*{0.2in}

\begin{flushleft}
{\Large
\textbf\newline{Evaluating Large Language Models' Ability Using a Psychiatric Screening Tool Based on Metaphor and Sarcasm Scenarios}
}
\newline
\\
Hiromu Yakura\textsuperscript{1*}
\\
\bigskip
\textbf{1} Max--Planck Institute for Human Development, 14195 Berlin, Germany
\\
\bigskip
* yakura@mpib-berlin.mpg.de
\end{flushleft}

\section*{Abstract}
Metaphors and sarcasm are precious fruits of our highly evolved social communication skills.
However, children with the condition then known as Asperger syndrome are known to have difficulties in comprehending sarcasm, even if they possess adequate verbal IQs for understanding metaphors.
Accordingly, researchers had employed a screening test that assesses metaphor and sarcasm comprehension to distinguish Asperger syndrome from other conditions with similar external behaviors (\eg{} attention-deficit/hyperactivity disorder).
This study employs a standardized test to evaluate recent large language models' (LLMs) understanding of nuanced human communication.
The results indicate improved metaphor comprehension with increased model parameters; however, no similar improvement was observed for sarcasm comprehension.
Considering that a human's ability to grasp sarcasm has been associated with the amygdala, a pivotal cerebral region for emotional learning, a distinctive strategy for training LLMs would be imperative to imbue them with the ability in a cognitively grounded manner.

\section{Introduction}

Since Aristotle's time~\cite{Aristotle335, Semino2016}, figurative language---encompassing metaphors and irony---has been regarded as one of the most sophisticated aspects of human communication.
Its realm of influence extends beyond the confines of literature, permeating everyday discourse, where its role is profound.
For example, in~casual phone conversations, figurative language occurs roughly once every 90 words~\cite{Bavelas2008}.
Hence, the~ability to decipher figurative language is considered a pivotal objective in linguistic development~\cite{Willinger2017}.

However, children possessing specific traits are recognized to grapple with challenges in comprehending such figurative language~\cite{Chahboun2021, Mention2024}, as~evidenced by studies suggesting distinct language processing patterns~\cite{Gold2010, Vulchanova2019}.
Particularly, the~comprehension of sarcasm in relation to Autism Spectrum Disorder (ASD) and other developmental disorders has received significant attention~\cite{Happ1995, Kalandadze2016, Fanari2023}.
Some researchers have associated this with the point that previous studies comparing children with ASD to typically developing peers often did not consider matching groups based on general language skills, suggesting that observed difficulties in understanding sarcasm might stem from general linguistic abilities~\cite{Gernsbacher2012, Fuchs2023}.
However, it is also reported that individuals that possessed the linguistic IQ to comprehend metaphors still struggled with grasping sarcasm~\cite{Adachi2004, Adachi2006}.
While its precise etiology remains elusive, individuals with the condition previously known as Asperger syndrome\footnote{While Asperger syndrome has been subsumed under the broader category of Social Pragmatic Communication Disorder in the DSM-5 \cite{APA2013}, this paper uses the term in its original context as it pertains to the MSST. This is to maintain coherence with the terminology used in MSST, given that our focus is on analyzing the behavior of LLMs rather than diagnosing human conditions.} are known to manifest reduced connectivity in the amygdala, the~cerebral region governing emotions~\cite{BaronCohen2000, Wang2023}.
Considering that such compromised connectivity of the amygdala impacts the cultivation of the \textit{theory of mind}~\cite{Heyes2014} in individuals with Asperger syndrome~\cite{BaronCohen1999}, Adachi~\etal~\cite{Adachi2006} postulated a potential commonality in the impediments to sarcasm comprehension.
They subsequently proposed a screening test, the~Metaphor and Sarcasm Scenario Test (MSST)~\cite{Adachi2004, Adachi2006}, to~distinguish children with the condition then referred to as Asperger syndrome from those with akin symptoms, such as attention-deficit/hyperactivity disorder and high-functioning autism.
This test measures comprehension of metaphors and sarcasm using five targeted questions for each category.
The resultant scores were leveraged to delineate a distinct subgroup of children who were likely to be diagnosed with Asperger syndrome, portraying a substantial disparity between metaphoric and sarcasm~comprehension.

Meanwhile, burgeoning advancements in large language models (LLMs), such as ChatGPT~\cite{OpenAI2022}, suggest an escalating potential for these models to undertake communications with humans~\cite{journals/tmlr/WeiTBRZBYBZMCHVLDF22}.
Under this trajectory, a~precise understanding of these models' limitations is deemed crucial not only for facilitating their communications with humans, but also for guiding their further development.
Accordingly, this study utilizes the MSST to scrutinize the aptitude of recent LLMs in comprehending metaphors and~sarcasm.

\subsection{Previous~Research}

The elucidation of the emergent capabilities of LLMs~\cite{journals/tmlr/WeiTBRZBYBZMCHVLDF22} has enthralled the attention of numerous scholars.
While much research has focused on assessing the extent to which LLMs can supplant human intellectual tasks~\cite{journals/corr/abs-2301-07597, journals/corr/abs-2303-10130, journals/corr/abs-2303-12712, Kung2023, Atarere2024, Kim2024}, such as medical decision making~\cite{Kung2023, Atarere2024, Kim2024}, several studies explored the depth of LLMs' grasp of human emotions and nuanced communications~\cite{journals/corr/abs-2212-05206, journals/corr/abs-2303-13988, journals/corr/abs-2307-00184, conf/emnlp/NematzadehBGGG18, journals/corr/abs-2209-01515, journals/corr/abs-2302-08399, Kosinski2023,10.3389/frobt.2023.1189525, conf/acl/AghazadehFY22, Loconte2023}.
Particularly, in~studies probing their ``theory of mind'' through standardized tasks~\cite{conf/emnlp/NematzadehBGGG18,journals/corr/abs-2209-01515,journals/corr/abs-2302-08399,Kosinski2023,10.3389/frobt.2023.1189525}, Kosinski~\cite{Kosinski2023} reported that GPT-4~\cite{OpenAI2023} exhibits abilities akin to those of an 8-year-old human.
However, research focusing on metaphors has been scant~\cite{conf/acl/AghazadehFY22, Loconte2023, Ichien2023, DiStefano2024}, with~only a few reports highlighting their ``superior'' performance to humans~\cite{Loconte2023, Ichien2023} or their potential role as an automated assessor in creativity scoring~\cite{DiStefano2024}, and~does not incorporate a comparative aspect with~sarcasm.

On the other hand, research efforts from an engineering perspective have addressed the enhancement of metaphorical and sarcasm comprehension in language models~\cite{conf/acl-figlang/SuFHLWC20, journals/nca/PotamiasSS20, Lal2022, journals/corr/abs-2107-02276, journals/corr/abs-2209-08141, conf/emnlp/ChakrabartySGM22, journals/corr/abs-2401-07078, conf/naacl/LiuCZN22}.
For instance, proposals have been posited to train dedicated models to detect metaphors~\cite{conf/acl-figlang/SuFHLWC20} and sarcasm~\cite{journals/nca/PotamiasSS20, journals/corr/abs-2107-02276} within text and explain them~\cite{Lal2022}.
However, these approaches are crafted to leverage the semantical proximity between input texts and instances of metaphors and sarcasm in training datasets, thereby raising concerns about their applicability to novel metaphors and sarcasm.
Alternatively, Prystawski~\etal~\cite{journals/corr/abs-2209-08141} proposed a method to improve metaphor comprehension in LLMs by refining the instructions provided to the model (\ie{} prompt).
In addition, some datasets and benchmark tasks~\cite{conf/emnlp/ChakrabartySGM22, journals/corr/abs-2401-07078, conf/naacl/LiuCZN22} have been released to encourage the development of LLMs that are capable of comprehending figurative language.
However, these efforts have not specifically focused on sarcasm comprehension, despite the knowledge from developmental psychology that revealed the contrastive distinction between metaphor and sarcasm understanding~\cite{Adachi2004, Adachi2006}.
Therefore, this study evaluates LLMs specifically on sarcasm comprehension to enable discussions that stand on the intricacies of human language~processing.

\subsection{Research~Questions}

This study aims to deepen the understanding of LLMs' capabilities in comprehending different forms of figurative language, specifically metaphors and sarcasm.
To achieve this, the~following research questions guide the investigation:
\begin{description}
    \item[RQ1] How do current large language models (LLMs) compare in their ability to comprehend metaphors versus sarcasm?
    \item[RQ2] How does the ability to comprehend metaphors and sarcasm vary among LLMs with different parameter counts?
\end{description}

By addressing these questions, this study seeks to uncover specific challenges and limitations that LLMs face in processing figurative language.
Furthermore, insights from developmental psychology will be integrated to highlight the cognitive distinctions between metaphor and sarcasm comprehension.
Ultimately, this study aims to guide future advancements in model training and development to enhance their~capabilities.

\subsection{Significance of The~Topic}

Considering the eminence of figurative language in human communication, assessing LLMs' comprehension abilities in this regard, particularly encompassing sarcasm, assumes paramount significance in augmenting their potential applications.
Furthermore, the~use of the standardized test for screening Asperger syndrome is substantive.
Within this test, children without intellectual disability\footnote{While we recognize the potential outdatedness and inadvisability of this term, for~reference, we note that Adachi~\etal~\cite{Adachi2004} used the label ``children without mental retardation'' in their results.} can attain high scores in both metaphorical and sarcasm comprehension~\cite{Adachi2004}.
On the other hand, the~development of linguistic intelligence alone leads to the attainment of metaphorical comprehension, but not sarcasm~\cite{Adachi2006}.
This distinction would provide a new clue to understanding the evolution of LLMs.
As Hagendorff~\cite{journals/corr/abs-2303-13988} alluded, we anticipate that a broader utilization of tools from developmental psychology elucidates the limitations and avenues for enhancement of~LLMs.

\section{Methods}
\unskip

\subsection{Material}

As mentioned above, the~MSST was employed to assess LLMs' comprehension of both metaphors and sarcasm.
Gibbs~\cite{Gibbs1994, Gibbs2011} stated that metaphor is a figure of speech where a word or phrase for a concept is used outside of its normal conventional meaning, often to express a similar concept.
Here, metaphors can be categorized into conventional ones, which are commonly used and widely understood within a culture (e.g., ``Love is a journey''), and~novel ones, which involve unique or less common associations that challenge the individual to forge new semantic links (e.g., ``We are driving in the fast lane on the freeway of love'').
The MSST covers both types of metaphors to incorporate some challenging~cases.

Irony, and~more specifically sarcasm, is another complex form of figurative language.
While irony involves expressing something contrary to what is meant, sarcasm often has a mocking or contemptuous tone~\cite{Garmendia2018}.
According to~\cite{Sperber1981IronyAT, Wilson_Sperber_Wilson_Sperber_2012}, ironic speakers express their dissociative attitude towards an utterance or thought they are echoing, rather than simply presenting an attitude dissociative from the content.
Meanwhile, Clark and Gerrig~\cite{Clark1984} describe sarcasm as a form of pretense, where the speaker pretends to be an injudicious person, expecting the listener to recognize the insincerity.
Given these points, understanding sarcasm requires inferring the speaker's intention or judgment, which is attributable to the theory of mind, and~thus can be challenging for children with the condition previously known as Asperger syndrome.
From this point, the~MSST employs scenarios that focus on this kind of sarcasm, as~presented in Figure~\ref{fig:example}.
We expected that using this test, grounded in figurative language studies, would enhance our understanding of LLM behavior relative to previous research on human~communication.

\begin{figure}[!ht]
\begin{boxedminipage}{0.92\textwidth}
The old man's garden is completely unkempt and overgrown. When an old lady went into the garden, she remarked, ``oh, what a beautiful garden it is!'' \\
\\
The old lady thinks the old man's garden~is... \\
 (a) tidy.       \\
 (b) lovely.     \\
 (c) messy.      \\
 (d) spacious.   \\
 (e) Don't know.
\end{boxedminipage}
\caption{The actual content of Q8 (taken from Adachi~\etal~\cite{Adachi2004}). No LLM selected the right choice, \ie{} (c).}
\label{fig:example}
\end{figure}
\unskip

\subsection{Procedure}

\begin{figure}[!ht]
\begin{adjustwidth}{-\extralength}{0cm}
\centering
\includegraphics[width=0.9\linewidth]{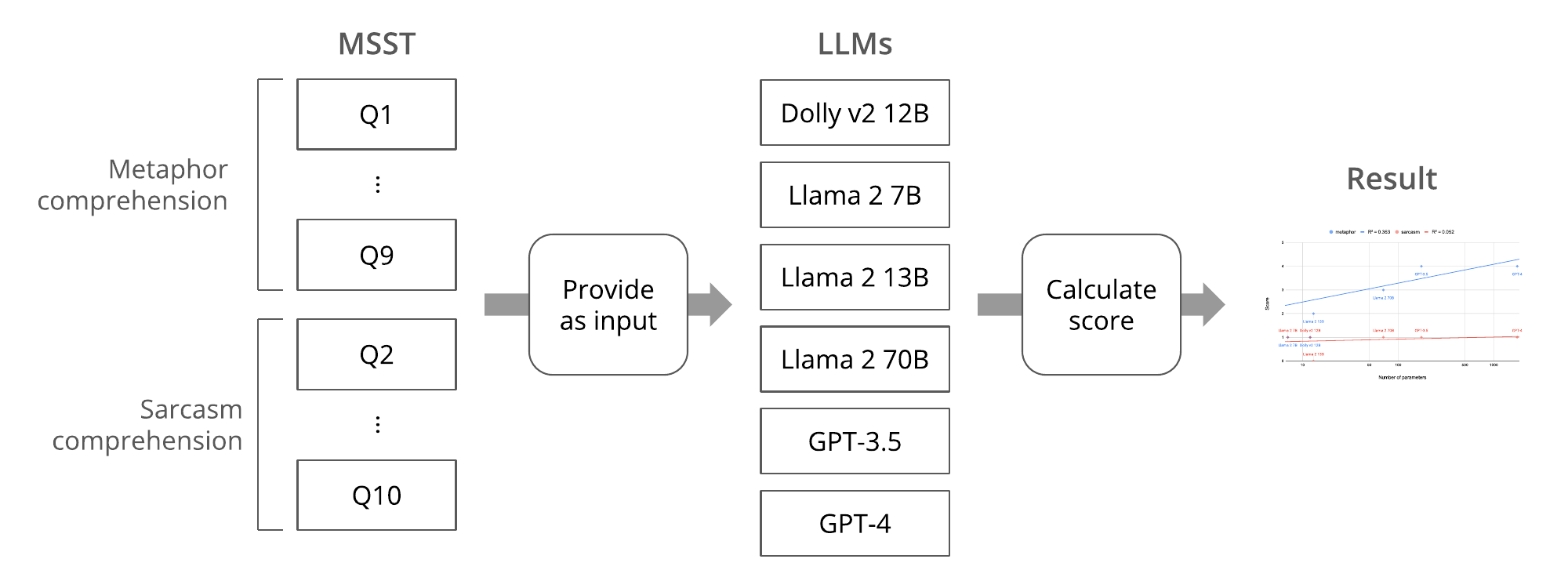}
\end{adjustwidth}
\caption{The procedure of this~study.}
\label{fig:procedure}
\end{figure}

The procedure of this study is illustrated in Figure~\ref{fig:procedure}.
To juxtapose with previous research on the theory of mind~\cite{Kosinski2023} and evaluate capacity variations based on differences in parameter count, we compared six LLMs, encompassing GPT-3.5 and GPT-4.
The others were selected based on the availability of public access to the models and their performance with instruction-following tuning, namely Dolly 2.0~\cite{DatabricksBlog2023DollyV2} and Llama 2 7B, 13B, and~30B~\cite{journals/corr/abs-2307-09288}.
For each model, we presented the MSST under the identical settings to conducting for human and calculated their correct answer rates.
For each question, we provided the models with an instruction sentence and an explanation of the context and offered five choices, from~which one had to be selected.
Each correct response is counted as a point, with~a maximum score of 5 for metaphor and sarcasm comprehension.
Notably, no human participant was engaged in this study because Adachi~\etal~\cite{Adachi2004} provided the scores of actual children, which enables us to compare their ability with those of the LLMs.
Also, the~source code necessary for replication, encompassing model version specifications, is publicly accessible at \href{https://doi.org/10.5281/zenodo.10981763}{doi:10.5281/zenodo.10981763}.

\section{Results}

\begin{figure}[!ht]
\centering
\includegraphics[width=\linewidth]{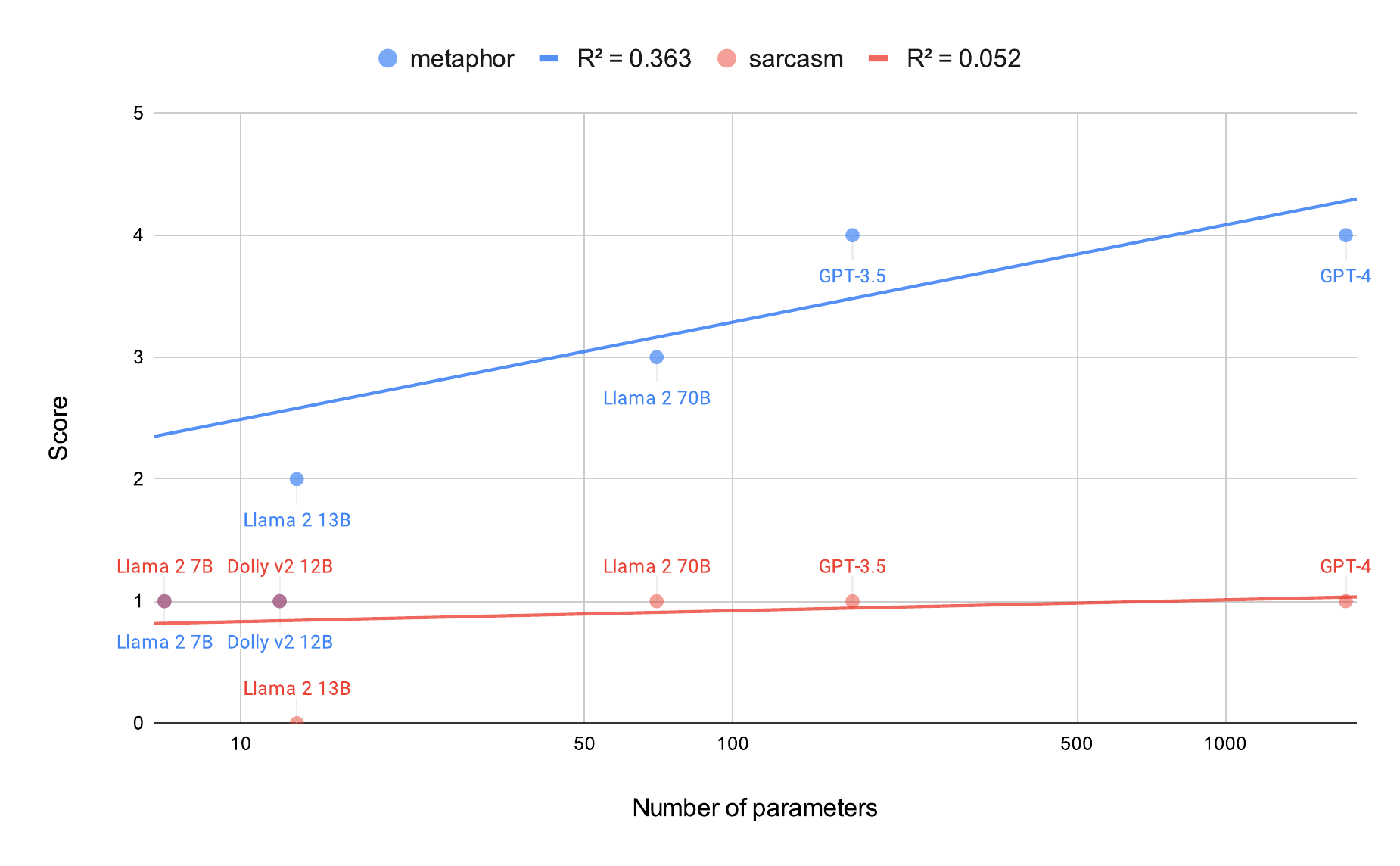}
\caption{The comparison of the scores of LLMs on the MSST and the number of their parameters. Note that the parameter count of GPT-4 is not officially announced and is based on an online article~\cite{Schreiner2023}.}
\label{fig:result}
\end{figure}

The results are presented in \tabref{tab:result} and Figure~\ref{fig:result}.
In terms of metaphor comprehension, it was observed that scores increased with an escalation in the number of parameters of the LLMs.
Here, Adachi~\etal~\cite{Adachi2004} reported a mean score of 4.1 for 199 children aged 8--10 years classified as without intellectual disability.
In this context, we can conclude that GPT-3.5 and GPT-4 exhibited the ability to grasp metaphors that are equivalent to those of 8 to 10-year-old children.
Note that this result corroborates Kosinski's assessment of GPT-4's inference capability based on the theory of mind~\cite{Kosinski2023}, which posited an aptitude akin to that of an 8-year-old.
In addition, this is consistent with the report from Sravanthi~\etal~\cite{journals/corr/abs-2401-07078}, in~which LLMs exhibited low performance for sarcasm understanding among various language understanding tasks, especially in comparison to human~performance.

\begin{table}[!ht]
\caption{The performances of LLMs on the MSST, in~which the metaphoric scenarios are odd-numbered and the sarcastic scenarios~even.}
\label{tab:result}
\begin{adjustwidth}{-\extralength}{0cm}
\centering
\begin{tabularx}{0.95\linewidth}{m{4cm}<{\raggedright}m{1.5cm}<{\centering}m{1.5cm}<{\centering}cccccccccc}
\toprule
\multicolumn{1}{c}{\multirow{2.4}{*}{\textbf{Model}}} & \multirow{2.4}{*}{\shortstack{\textbf{Metaphor}\\\textbf{Score}}} & \multirow{2.4}{*}{\shortstack{\textbf{Sarcasm}\\\textbf{Score}}} & \multicolumn{10}{c}{\textbf{Questions}} \\
\cmidrule{4-13}
              &   &   & \textbf{Q1}         & \textbf{Q2}         & \textbf{Q3}         & \textbf{Q4} & \textbf{Q5}         & \textbf{Q6        } & \textbf{Q7}         & \textbf{Q8} & \textbf{Q9        } & \textbf{Q10 }       \\
\midrule
Dolly v2 12B  & 1 & 1 &            &            & \checkmark &    &            &            &            &    &            & \checkmark \\
Llama 2 7B    & 1 & 1 &            &            & \checkmark &    &            & \checkmark &            &    &            &            \\
Llama 2 13B   & 2 & 0 & \checkmark &            &            &    &            &            &            &    & \checkmark &            \\
Llama 2 70B   & 3 & 1 & \checkmark &            & \checkmark &    &            & \checkmark & \checkmark &    &            &            \\
GPT-3.5       & 4 & 1 & \checkmark &            & \checkmark &    & \checkmark & \checkmark & \checkmark &    &            &            \\
GPT-4         & 4 & 1 &            & \checkmark & \checkmark &    & \checkmark &            & \checkmark &    & \checkmark &            \\
\midrule
Children w/o  & \multirow{2}{*}{4.1} & \multirow{2}{*}{3.3} & \multicolumn{10}{c}{\multirow{2}{*}{---}} \\
intellectual disability \cite{Adachi2004} \\
\bottomrule
\end{tabularx}
\end{adjustwidth}
\end{table}
\unskip

Conversely, with~regard to sarcasm comprehension, none of the LLMs managed to answer more than a single question correctly, as~shown in the even-numbered questions in \tabref{tab:result}.
For instance, none of the LLMs could provide a correct response to Q8\footnote{The content of all questions is available in the source code published at \href{https://doi.org/10.5281/zenodo.10981763}{doi:10.5281/zenodo.10981763}.} (see Figure~\ref{fig:example}).
We may infer that it would not pose an incredibly daunting challenge for individuals possessing a fundamental conception of sarcasm, given that the average score for the 199 children was 3.3~\cite{Adachi2004}.
However, both GPT-3.5 and GPT-4 opted for ``(b) lovely,'' yielding a score lower than the average of 1.8 achieved by 66 children diagnosed with Asperger syndrome~\cite{Adachi2006}.

Here, since the MSST is published as a part of the existing paper~\cite{Adachi2004}, we recognize the possibility that its questions might have been included in the LLMs' training data, as~other benchmark datasets have experienced~\cite{conf/aaai/LiF24, journals/corr/abs-2308-08493}.
However, if~this were the case, it would be unusual for the models to comprehend only the metaphoric scenarios while scoring poorly on the sarcastic ones, suggesting that the questions might have been unfamiliar to the models.
Furthermore, even assuming the questions were part of the training data, the~results interestingly suggest that the models face difficulties in processing sarcasm, although~they can effectively memorize metaphors.
This underlines the apparent difficulty these models faced in comprehending sarcasm, irrespective of whether the MSST questions were included in their training~data.

\section{Discussion}

Our investigation revealed that, whereas recent LLMs have substantially heightened their capacity for metaphorical comprehension with increased parameters, they still grapple with sarcasm comprehension.
For humans, such a disparity is a distinctive trait associated primarily with individuals with the condition previously referred to as Asperger syndrome, and~various explanations have been proposed, focusing on the distinct cognitive attributes characterizing this group~\cite{Loukusa2009}.

One explanation highlights that, while metaphor comprehension is driven by linguistic intelligence, sarcasm comprehension necessitates emotional intelligence, which involves recognizing and interpreting emotions and social cues~\cite{Adachi2006, Adachi2004}.
Metaphors can often be understood by recognizing the discordance between their content and factual reality, which relies on cognitive processes that develop relatively early in life.
For instance, children can understand basic metaphors by the age of three~\cite{DiPaola2020, Pouscoulous2020}.
In contrast, as~previously mentioned, sarcasm often involves content that remains plausible and requires the listener to detect the incongruity between the speaker's literal words and their actual intent, which is inferred from the context~\cite{Sperber1981IronyAT, Wilson_Sperber_Wilson_Sperber_2012, Clark1984}.
This process involves the higher-order theory-of-mind ability to understand that others have beliefs, desires, and~intentions different from one's own~\cite{Happ1993}.
Specifically, Mazzarella and Pouscoulous~\cite{Mazzarella2021, Mazzarella2023} suggest that sarcasm comprehension depends on the emergence of vigilance towards deception, specifically ``second-order epistemic vigilance,'' which refers to the ability to assess others' abilities to scrutinize deception.
This explains why even children with typical development struggle with sarcasm comprehension until at least six years old~\cite{Filippova2008, Pexman2023}.
Consequently, this point presents a formidable challenge for individuals with the condition akin to Asperger syndrome, who often exhibit less performance in inferring others' beliefs in the theory of mind tasks~\cite{BaronCohen1999, BaronCohen2000, Wang2023}.

Another explanation is rooted in the concept of weak central coherence~\cite{Frith1998, Frith2008}, which is prevalent in individuals with the condition previously referred to as Asperger syndrome~\cite{Jolliffe2000, LeSournBissaoui2011}.
They are said to exhibit a tendency to process information locally rather than globally, struggling to integrate information from multiple sources to derive context-dependent meanings~\cite{Frith1998, Frith2008}.
In other words, their cognitive processes are significantly influenced by the literal meanings of certain words or expressions from isolated segments of communication, hindering their comprehension of sarcasm.
These discussions provide inspiration for devising novel approaches to enhance the performance of LLMs, particularly their capacity to understand~sarcasm.

For example, while the current training method of LLMs being reliant on extensive text datasets is effective in improving their linguistic intelligence, it may be inadequate when fostering emotional intelligence.
In such a scenario, supplementing the current training regime with a distinct kind of training data could amplify the acquisition, akin to how individuals who were diagnosed with Asperger syndrome resolve social challenges a posteriori through social skill training~\cite{Rao2007}.
However, given the scarcity of available scenarios of social skill training compared to the datasets used for LLM training, strategic efforts are requisite to construct novel large-scale datasets or adopt new data-efficient fine-tuning techniques like Low-Rank Adaptation (LoRA)~\cite{conf/iclr/HuSWALWWC22}.
In addition, given that multimodal pre-training using visual information is effective for knowledge transfer in LLMs~\cite{conf/mm/MuraokaBMBLZ23}, constructing multimodal datasets that include facial expressions may also prove effective, as~it can enable capturing nuanced human emotion via pre-training as a vision large language~model.

Furthermore, our findings can be associated with the fact that the majority of recent LLMs, including those examined in this study, are obtained using instruction-based tuning methods, such as InstructGPT~\cite{conf/nips/Ouyang0JAWMZASR22}.
The popularity of these tuning methods stems from their effectiveness in enabling LLMs to follow user instructions accurately and avoid biased responses.
However, it is conceivable that such tuning processes might impede the models' ability to gauge human subjective judgments---a critical aspect in comprehending sarcasm, as~highlighted earlier.
This is because bias inherently intertwines with subjective judgments~\cite{Gilovich2002}, and~suppressing potentially biased outputs would inadvertently constrain inferences pertinent to human subjective judgment.
From these points, it would be imperative to explore prudent strategies that enable LLMs to infer human subjective judgment while restraining biased outputs.
Addressing this delicate balance can facilitate LLMs in engaging users through nuanced figurative language in witty~dialogues.

Notably, it is known that LLMs are susceptible to misdirection by unrelated contextual information provided in a small portion of prompts~\cite{pmlr-v202-shi23a}.
This raises the intriguing possibility that cognitive processes akin to those observed in individuals who were diagnosed with Asperger syndrome might influence their ability to comprehend sarcasm.
Nevertheless, as~a limitation of this study, it is imperative to consider that the behavior of LLMs may not directly mirror human neurological functions, even though high-level commonalities have been observed~\cite{conf/acl/EvansonLK23, Caucheteux2023}.
We also need to mention that our results did not involve statistical tests due to the difficulty of forming a participant pool with LLMs and can be updated with future advancements in LLMs.
Nonetheless, we believe that insights cultivated in developmental psychology can play a pivotal role in unraveling the behavior of LLMs, as~exemplified by this~study.

\section{Conclusions}

This study aimed to evaluate the capabilities of LLMs in comprehending figurative language, specifically metaphors and sarcasm.
Specifically, we used the MSST and revealed that while LLMs show improved performance in metaphor comprehension with an increase in model parameters, their ability to comprehend sarcasm remains significantly limited.
Our approach of using the established screening test allowed us to find the alignment of this disparity with insights from developmental psychology.
Based on this, this study discussed the potential importance of emotional and social context in training methodologies, highlighting the importance of focusing on sarcasm comprehension in LLMs to better understand their limitations and guide future advancements.
Future research should continue to develop strategies to enhance the emotional intelligence of LLMs, ultimately improving their application possibility within the full spectrum of human~communication.

\vspace{6pt} 


\section*{Acknowledgements}

This work was supported in part by JSPS KAKENHI (JP21J20353) and JST ACT-X (JPMJAX200R), Japan.

\begin{adjustwidth}{-\extralength}{0cm}

\end{adjustwidth}
\end{document}